# What is the Best Grid-Map for Self-Driving Cars Localization? An Evaluation under Diverse Types of Illumination, Traffic, and Environment

Filipe Mutz[1,2,*], Thiago Oliveira-Santos[2], Avelino Forechi[3], Karin S. Komati[1], Claudine Badue[2], Felipe M. G. França[4], and Alberto F. De Souza[2]

**ABSTRACT**

The localization of self-driving cars is needed for several tasks such as keeping maps updated, tracking objects, and planning. Localization algorithms often take advantage of maps for estimating the car pose. Since maintaining and using several maps is computationally expensive, it is important to analyze which type of map is more adequate for each application. In this work, we provide data for such analysis by comparing the accuracy of a particle filter localization when using occupancy, reflectivity, color, or semantic grid maps. To the best of our knowledge, such evaluation is missing in the literature. For building semantic and colour grid maps, point clouds from a Light Detection and Ranging (LiDAR) sensor are fused with images captured by a front-facing camera. Semantic information is extracted from images with a deep neural network. Experiments are performed in varied environments, under diverse conditions of illumination and traffic. Results show that occupancy grid maps lead to more accurate localization, followed by reflectivity grid maps. In most scenarios, the localization with semantic grid maps kept the position tracking without catastrophic losses, but with errors from 2 to 3 times bigger than the previous. Colour grid maps led to inaccurate and unstable localization even using a robust metric, the entropy correlation coefficient, for comparing online data and the map.

*Keywords*: Robotics, Self-Driving Cars, Localization, Mapping, Grid Maps.

## I. INTRODUCTION

Self-driving cars have potential to change several aspects of society, ranging from the economy to city organization and lifestyle. If the technology succeeds to be adopted, jobs, businesses, and even climate may be affected in the long term (National Highway Traffic Safety Administration, 2019; Securing America's Future Energy, 2019; United States Department of Labor, 2019). SAE International (formerly simply SAE, or Society of Automotive Engineers) published a classification system for assessing levels of autonomy of self-driving cars (SAE International, 2016). According to their system, autonomy can range from 0 (no automation) to 5 (full autonomy). In order to achieve level 5, self-driving cars must be able to operate in complex dynamic environments such as city centers, residential areas, and suburbs.

For driving safely in these environments, self-driving cars have to perceive the environment and its rules. Unfortunately, obtaining this information directly from sensors may be unfeasible. Sensors are subject to noise, failures and, currently, cannot observe the entirety of the environment in all of its details. In addition, existence of proper traffic infrastructure, e.g., clear unambiguous traffic signs, traffic lights, lane marks, and pavement marks, cannot be guaranteed. Finally, finding relevant information in raw sensor data can be a computationally expensive task, even with perfect sensors and infra-structure. Since the size and costs of hardware can be impacted by the computational requirements of software, companies are likely to prefer solutions with lower computational demands. Maps can store and integrate information extracted from sensors, such as the layout of roads, their components and rules. They can also be built before autonomous operation with the aid of human experts. Therefore, maps are commonly used as alterative for extracting information from sensors online.

---

[*] Corresponding author: Filipe Mutz, Coordenadoria de Sistemas de Informação, Instituto Federal de Educação, Ciência e Tecnologia do Espírito Santo (IFES), Serra, ES, 29173-087, Brazil.
[1] Instituto Federal de Educação, Ciência e Tecnologia do Espírito Santo (IFES), Serra, ES, 29173-087, Brazil.
[2] Departamento de Informática, Universidade Federal do Espírito Santo (UFES), Vitória, ES, 29075-910, Brazil.
[3] Instituto Federal de Educação, Ciência e Tecnologia do Espírito Santo (IFES), Aracruz, ES, 29192-733, Brazil.
[4] Universidade Federal do Rio de Janeiro (UFRJ), Rua Horácio Macedo, Bloco G, 2030 - 101 - Cidade Universitária da Universidade Federal do Rio de Janeiro, Rio de Janeiro - RJ, 21941-450, Brazil.
E-mail Adresses: filipe.mutz@ifes.edu.br (Filipe Mutz), todsantos@lcad.inf.ufes.br (Thiago Oliveira-Santos), avelino.forechi@ifes.edu.br (Avelino Forechi), kkomati@ifes.edu.br (Karin S. Komati), claudine@lcad.inf.ufes.br (Claudine Badue), felipe@cos.ufrj.br (Felipe M. G. França), alberto@lcad.inf.ufes.br (Alberto F. De Souza).

For using and updating maps, as well as for other tasks, such as object tracking, and planning, the localization of the self-driving car has to be known. Localization algorithms often take advantage of maps for estimating poses, with grid maps being the most frequently used map in self-driving cars (Badue, et al., 2020). Each type of grid map has positive and negative features. Reflectivity grid maps (Badue, et al., 2020; Levinson & Thrun, 2010; Wolcott & Eustice, 2017; Levinson, Montemerlo, & Thrun, 2007) and colour grid maps (Hornung, Wurm, Bennewitz, Stachniss, & Burgard, 2013; Bârsan, Liu, Pollefeys, & Geiger, 2018) store rich image-like representations of the environment, while occupancy grid maps (Thrun, Burgard, & Fox, 2005) and semantic grid maps (Kostavelis & Gasteratos, 2015; Yang, Huang, & Scherer, 2017) contain information that can be used for both localization and planning. Figure 1 (b)-(e) present examples of these maps for the region in Figure 1 (a). Semantic grid maps can also be used in other tasks, e.g., removing trails of moving objects from maps (Guidolini, Carneiro, Badue, Oliveira-Santos, & De Souza, 2019).

Maintaining and using several maps is computationally expensive in terms of processing, memory and network bandwidth (if maps are not stored in the car). The comparison of maps in diverse tasks and scenarios may provide information for decision makers to choose the most adequate map for each application. To the best of our knowledge, such evaluation is missing in the literature.

In this work, we evaluate the impact of using occupancy, reflectivity, color, or semantic grid maps with regards to localization accuracy. Experiments were performed in daylight, nighttime and twilight, on free and busy streets, and in different environments, such as, a university campus, a neighborhood with tree lined streets, suburbs, avenues, and in an airport. Results show that occupancy grid and reflectivity grid maps lead to more accurate localization, with occupancy grid maps being the more precise. Localization with semantic grid maps kept the position tracking without losses in most scenarios, however with errors from 2 to 3 times bigger than the first two. Colour grid maps led to inconsistent and inaccurate localization even using a robust metric, the entropy correlation coefficient (Astola & Virtanen, 1983; Thanh, Kim, Hong, & Ngo Lam, 2018), for trying to reduce the impact of illumination when comparing online data with the map.

Hence, the main contributions of this work are:

- Evaluation of which type of grid map leads to more accurate localization. Experiments show that occupancy grid maps and reflectivity grid maps are the ones with smaller error.

- Demonstration that semantic grid maps can be successfully used for localization. Semantic information is extracted from images using the deep neural network DeepLabv3+ (Chen, Zhu, Papandreou, Schroff, & Adam, 2018) trained on the Cityscapes dataset (Cordts, et al., 2016).

- Evaluation of localization accuracy using the entropy correlation coefficient as a metric for comparing color data. Results show that even using this robust metric, localization with color data is inaccurate and unstable.

- Development of a two-step process based on GraphSLAM (Thrun, Burgard, & Fox, 2005; Thrun & Montemerlo, 2006) for estimating the car path and building maps. Previous techniques (Badue, et al., 2020; Mutz, et al., 2016) led to inaccurate loop closures in our datasets.

- A new technique for computing the ground truth and evaluating the localization. The technique is also based on GraphSLAM.

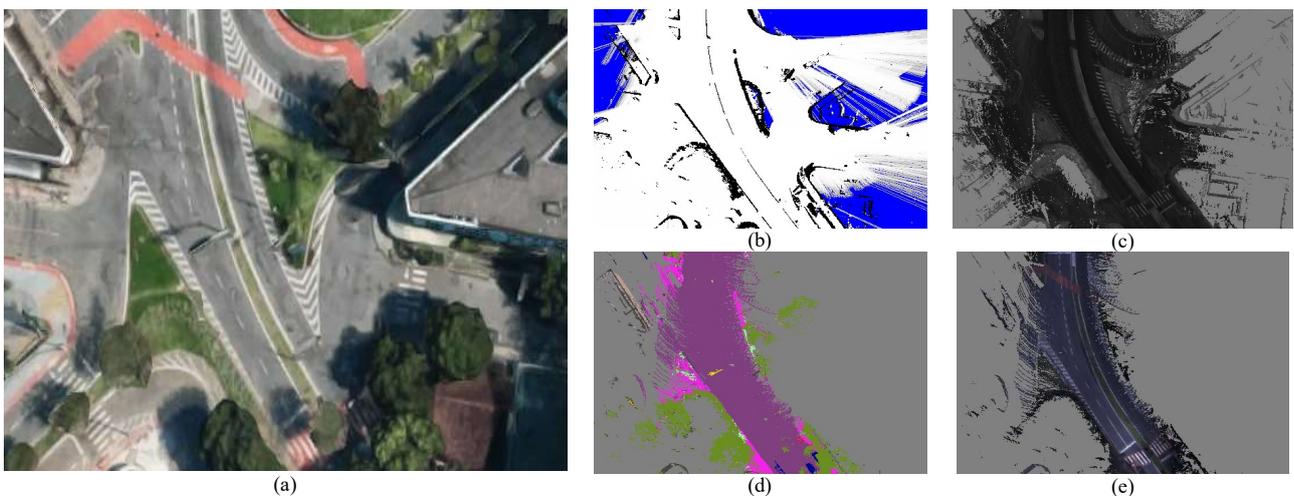

Figure 1. Different types of grid maps for the region presented in (a). (b) Occupancy grid map. (c) Reflectivity grid map. (d) Semantic grid map. (e) Colour grid map**.**

The remaining of the paper is organized as follows. Section II presents related works and discusses limitations of previous methods. Section III describes the hardware and software architectures of the self-driving car used in the experiments, as well as preprocessing applied to its sensors' data. Section IV and V present our key contributions, the systems for localization and mapping, respectively. Section VI describes the experimental evaluation and discusses the results. Finally, Section VII presents conclusions obtained from the experiments and lists opportunities of future works.

## II. Literature

This section discusses previous works regarding mapping and localization for self-driving cars. The section starts by presenting techniques based on simultaneous localization and mapping (SLAM) for estimating the poses of a robot given sensors' data. These poses can be used for building maps of the environment. Subsequently, it presents previous works on grid mapping with an emphasis on semantic and color grid maps. Finally, algorithms for estimating the localization of robots in relation to grid maps are discussed.

### A. Simultaneous Localization and Mapping (SLAM)

The problem of mapping unknown environments is usually addressed using probabilistic approaches which can be classified into online and offline. Online algorithms estimate at each step the pose of the self-driving car and the map. Offline SLAM algorithms, on the other hand, try and solve the Full-SLAM problem which consists of estimating all of the vehicle path and the map using all the data available (Badue, et al., 2020; Thrun, Burgard, & Fox, 2005; Durrant-Whyte & Bailey, 2006; Bailey & Durrant-Whyte, 2006).

Examples of online SLAM algorithm includes the ones based on Kalman filter such as the Extended Kalman Filter (Weingarten & Siegwart, 2005), the Unscented Kalman Filter (UKF) SLAM (Wan & Van Der Merwe, 2000), and Iterated Sigma Point Kalman Filter SLAM (Sibley, Sukhatme, & Matthies, 2006). Even being the standard solution to the SLAM problem for several years, the EKF-SLAM algorithm has drawbacks that prevent its use in outdoor and/or large-scale environments (Durrant-Whyte & Bailey, 2006; Bailey & Durrant-Whyte, 2006; Huang & Dissanayake, 2007; Montemerlo, Thrun, Koller, & Wegbreit, 2002; Montemerlo & Thrun, 2007).

Montemerlo et al. introduced two online SLAM algorithms, the FastSLAM (Montemerlo, Thrun, Koller, & Wegbreit, 2002) and the FastSLAM 2.0 (Montemerlo & Thrun, 2007). These algorithms use Monte Carlo Markov Chain techniques (particle filters) to factorize the SLAM posterior into a product of conditional map distributions and a distribution over robot paths. This factorization (called Rao–Blackwellization) enables the introduction of nonlinear motion models and notably speeds up the SLAM algorithm.

Thrun and Montemerlo (Thrun & Montemerlo, 2006) state that the main disadvantage of online SLAM algorithms is their inability to revisit sensors data, in particular for handling loop closures. Lu and Milios (Lu & Milios, 1997) introduced offline techniques for estimating the localization of a robot when performing SLAM. Their work as well as subsequent works (Duckett, Marsland, & Shapiro, 2000; Frese & Hirzinger, 2001; Konolige, 2004) showed that it is possible to increase the accuracy of the localization estimates by memorizing the data until the estimation process is complete and only then building the map. Golfarelli, Maio, and Rizzi (Golfarelli, Maio, & Rizzi, 1998) showed that the SLAM posterior probability can be modeled as a sparse graph. Moreover, the optimization of this sparse graph leads to a sum of nonlinear quadratic factors. When optimizing this objective function (the sum of nonlinear quadratic factors), it is possible to obtain the most likely map and poses given the sensors' data.

Thrun and Montemerlo (Thrun & Montemerlo, 2006) proposed the GraphSLAM algorithm which builds on top of these works. In the GraphSLAM, nodes represent model variables (the vehicle poses and the map), and edges represent sensor measurements and their covariances. In outdoor applications, the number of map variables (e.g., landmarks) tends to be large. Attempting to reduce the computational cost of the method, the authors propose to marginalize the map variables out of the posterior distribution. They do so by transforming the restrictions imposed by the map in relationships between poses. In particular, the paths of the vehicle in different visits to loop closure regions become dependent. These dependencies are modelled as edges in the graph. For adding these edges, the displacement between the paths must be estimated along with the covariance of the estimates.

Levinson, Montemerlo, and Thrun (Levinson, Montemerlo, & Thrun, 2007) propose to solve the displacement estimation problem using map-matching. They assume that the orientation of the vehicle is known, and estimate the displacement in the longitudinal and lateral directions. Although it is a valid procedure when the errors in the orientation measurements are negligible, the method would require a limiting amount of computation if the orientation has to be estimated. Levinson and Thrun (Levinson & Thrun, 2010), and posteriorly Mutz et al. (Mutz, et al., 2016) used the Generalized Iterative Closest Point algorithm (Segal, Haehnel, & Thrun, 2009) for estimating the poses of a vehicle in subsequent visits to loop closure regions.

### B. Grid Mapping

Given the sensors' data and the poses of the self-driving car, the mapping problem reduces to projecting the sensors' data into the map. The choice of which information to store and how to represent it depends on the application. For self-driving cars, the most common representation are grid maps (Badue, et al., 2020), in particular occupancy and reflectivity grid maps (Badue, et al., 2020; Levinson & Thrun, 2010; Wolcott & Eustice, 2017; Veronese, et al., 2016).

The creation of semantic grid maps has been an active area of research (Zhao & Chen, 2016; Vineet, et al., 2015; McCormac, Handa, Davison, & Leutenegger, 2017; Sengupta, Sturgess, Ladicky, & Torr, 2012). Zhao and Chen used RGB-D sensors for creating 3D semantic maps of indoor scenes (Zhao & Chen, 2016). They performed visual semantic segmentation of images using a support vector machine (SVM) classifier and dense conditional random fields (CRF). Vineet et al. used a stereo camera for creating 3D semantic grid maps of large-scale environments (Vineet, et al., 2015). The camera path is estimated using visual odometry. The visual data are projected in the map using the depth maps computed by stereo-matching algorithms. The authors claim that algorithms based on stereo cameras produce data that are noisier than LiDARs' measurements, but they are suitable for both, large robots, and wearable glasses/headsets.

McCormac et al. (McCormac, Handa, Davison, & Leutenegger, 2017) applied a similar approach posteriorly, but using convolutional neural networks (CNNs) for semantic segmentation, and the ElasticFusion algorithm for path estimation. Yang, Huang, and Scherer (Yang, Huang, & Scherer, 2017) used CNNs and CRFs for semantic segmentation of sensors' data which are integrated in 3D semantic grid maps. Similarly, Sengupta, Sturgess, and Torr (Sengupta, Sturgess, Ladicky, & Torr, 2012) used two CRFs to build 2D semantic grid maps of urban areas. The first one is used for instantaneous visual semantic segmentation, while the second one is used for aggregating the segmentations on the map. Neither of the previous works evaluates the feasibility of using the semantic grid maps for localization. It is worth noting, however, that semantic landmarks maps have already been used for localization (Atanasov, Zhu, Daniilidis, & Pappas, 2014; Atanasov, Zhu, Daniilidis, & Pappas, 2016; Ma, et al., 2019).

*C. Localization*

Poggenhans et al (Poggenhans, Salscheider, & Stiller, 2018) proposed a localization method that uses road markings and borders as grid map features for map matching and later fusion via UKF. Road markings are detected on accumulated top-view images created by projecting RGB-D point clouds into a grid map. They are detected using traditional image processing pipeline. On the other hand, road boarder detection is carried out using semantic segmentation based on ResNet (Wu, Shen, & van den Hengel, 2019) trained on the Cityscapes dataset. They define the road border as the transition from road to any other label. Similar to the road marking detection, the labelled image is transformed to an accumulated top-view grid map. Next, they employ map matching to find associations between elements in the map and the detected features. They fuse the candidate poses using UKF in order to select the pose with the highest observation likelihood, then update the filter with it and discard the others. Their results showed that the localization method has higher accuracy (8 cm and 19 cm in y and x directions, respectively) in typical road scenarios compared to narrow streets with an average lateral and longitudinal error of 58 cm and 37 cm, respectively, in different weather conditions such as sunny, cloudy and rain.

Veronese et al. (Veronese, et al., 2015) proposed a localization method that compares color satellite maps with re-emission maps. A particle filter algorithm is employed to estimate the car's pose. The particles' weights are given by the normalized mutual information (NMI) between instantaneous remission maps and the satellite maps. The method was evaluated on a 6.5 km dataset and it achieved an accuracy of 89 cm. One advantage of this method is that it does not require building a map specifically for the method.

Berger et al. (Berger, Orf, Muffert, & Zollner, 2018) proposed a direct registration of geometrical features (e.g. lane markings, curbs and poles) to grid maps for localization. The registration is done in real-time using GPUs running exhaustive search in combination with a multi-level-resolution search. Then, a Kalman Filter (KF) fuses the registration results with odometry information in order to localize the vehicle with an accuracy of 10 cm and 18 cm in y and x directions, respectively and is also robust against challenging situations like difficult maneuvers or missing localization features in the map.

Veronese et al. (Veronese, et al., 2016) proposed a localization method based on particle filters that computes the particles' likelihood by matching 2D instantaneous occupancy grid-maps and 2D offline occupancy grid-maps. Two map-matching distance functions were evaluated: an improved version of the traditional Likelihood Field distance between two grid-maps, and an adapted standard cosine distance between two high-dimensional vectors. An experimental evaluation on the IARA self-driving car demonstrated that the localization method can operate at about 100 Hz using the cosine distance function, with lateral and longitudinal errors of 13 cm and 26 cm, respectively.

Wolcott and Eustice (Wolcott & Eustice, 2017) proposed a probabilistic localization method that models the world as a multiresolution grid map. Each cell of the map stores the parameters of mixtures of Gaussians representing the height and reflectivity of the environment. An extended Kalman filter (EKF) localization algorithm is used to estimate the car's pose by registering 3D point clouds against the multiresolution-maps. The method was evaluated on two driverless cars in adverse weather conditions and presented localization estimation errors of about 15 cm.

Levinson et al. (Levinson, Montemerlo, & Thrun, 2007) proposed a localization method that uses reflectivity grid maps for estimating the localization of self-driving cars. They used the multi-layer LiDAR Velodyne HDL-64E LIDAR. A 2-dimension histogram filter (Thrun, Burgard, & Fox, 2005) is employed to estimate the self-driving car position. Their method has shown a root mean squared (RMS) lateral error of 9 cm and a root mean squared error (RMS) longitudinal error of 12 cm.

Clemens et al. (Clemens, Kluth, & Reineking, 2019) proposed an extended representation of the uncertainty in the occupancy grid map. They modelled the occupancy probabilities of the grid cells as beta-distributed random variables. This is in contrast to the classical approach of a Bernoulli distribution per cell. Their approach allows to quantify the uncertainty and also distinguish between the different causes of uncertainty such as missing or conflicting information based on the Shannon entropy. They extended the Rao-Blackwellized Particle Filter (RBPF) to support the Beta distribution formalism and showed equivalent results in localization and map accuracy as the classical one. In addition, their approach allows classifying the true state of a cell with a higher accuracy and to detect dynamics in the environment. Their map uncertainty model can benefit navigation by introducing safer cost functions, active exploration by identifying regions with conflicting measures and map maintenance by helping to detect and remove dynamic obstacles.

Lu et al. (Lu, Zhou, Wan, Hou, & Song, 2019) proposed a Deep Learning approach to the localization problem using data from LiDAR sensors. Their proposed approach achieves centimeter-level accuracy, comparable to state-of-the-art methods based on Bayesian filters and feature engineering. They replaced this probabilistic framework with deep neural network architectures trained end-to-end, called L3-Net. L3-Net employs RNNs to model vehicle dynamics and 3D CNN to extract local descriptors from point clouds for later correspondence in different driving scenarios. They collected 380.5 km of sensor data during multiple trials over six different routes in San Francisco (California) to evaluate the effectiveness of their approach. They found in a unseen route, 37.7 km long, superior or equivalent accuracy compared to Levinson and Thrun (Levinson & Thrun, 2010) and others. Their method achieved generalization with a RMS lateral error of 5.5 cm and a RMS longitudinal error of 3.7 cm.

### III. THE ROBOTIC PLATFORM

This section presents the self-driving car used in this work, along with its sensors and actuators. The section also describes the preprocessing applied to sensors' data.

*A. Hardware*

The Intelligent Autonomous Robotic Automobile (IARA) is a research self-driving car developed at LCAD/UFES (the High-Performance Computing Laboratory of the Universidade Federal do Espírito Santo, Brazil). It was the first self-driving car to travel autonomously 74 km on urban roads and highways in Brazil.

IARA is based on a 2011 Ford Escape Hybrid adapted for autonomous operation by Torc Robotics (https://torc.ai/) and by LCAD's team. Sensors used in this work are a Velodyne HDL-32E LiDAR, a front-facing Bumblebee XB3 stereo camera, an Xsens MTi IMU, a Trimble RTK GPS, and an odometry sensor. For conciseness, these sensors will be referred as Velodyne, camera, IMU, GPS, and odometry from now on. The arrangement of the sensors in IARA as well as their coordinate systems are illustrated in Figure 2.

Data collected by sensors are as follows. GPS provides positioning information in relation to a global coordinate system on Earth. The odometry sensor measures the vehicle's linear velocity and steering wheel angle. IMU provides accelerations and angular velocities in the directions of the x, y, and z axes, as well as the sensor's orientation in relation to a global coordinate system on Earth. Velodyne is composed of 32 lasers assembled one above the other. Each laser measures the distance to the nearest object and its surface reflectivity. All 32 range and reflectivity readings are assumed to be obtained at the same time and they are referred as a shot. A rotating motor turns the 32 lasers around the vertical axis. This movement allows the sensor to provide a car-centered sphere-like 3D view of the environment. For each shot, the rotating motor's horizontal angle is also given. The stereo camera system contains three cameras shifted horizontally by 12cm and 24cm. Only images from the right camera are used in this work. Figure 3 exemplifies images and Velodyne point clouds in different environments.

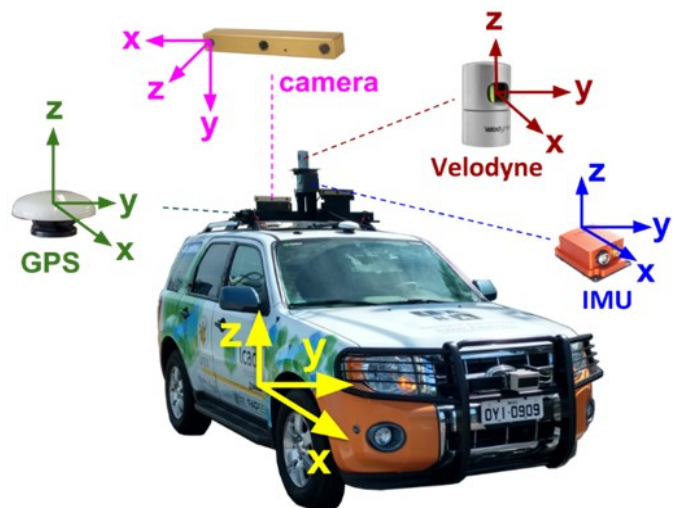

Figure 2. Sensors used for mapping and localization, their coordinate systems, and their arrangement in the Intelligent Autonomous Robotic Automobile (IARA) self-driving car.

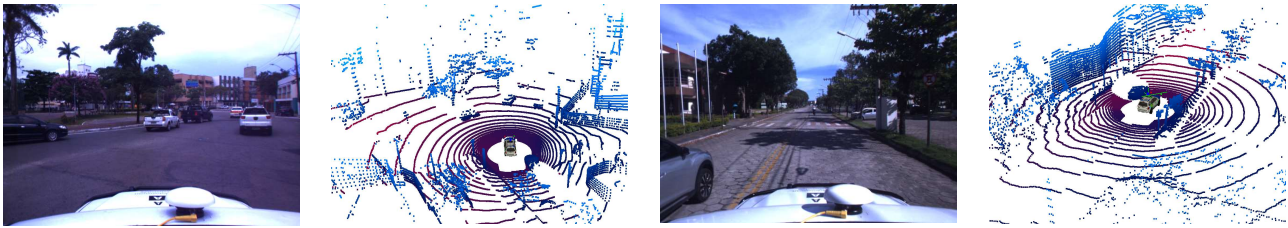

Figure 3. Examples of camera images and Velodyne point clouds in two environments.

*B. Software*

IARA's software architecture is based on the Carnegie Mellon navigation (CARMEN) toolkit (Montemerlo, Roy, & Thrun, 2003) and it is available at https://github.com/LCAD-UFES/carmen_lcad as open source. The g2o framework (Grisetti, Kümmerle, Strasdat, & Konolige, 2011) was employed for implementing the GraphSLAM algorithm. G2o provides tools for modeling probabilistic processes as graphs and efficient optimization methods for inference.

The proposed mapping and localization systems were implemented as modules in CARMEN. The map building module has two modes of operation, offline mode, and online mode. Offline mode is employed before autonomous operation for building maps of previously unseen environments. These offline maps represent static structures as realistically as possible as illustrated by the occupancy grid map in Figure 4 (a). Online mode is employed during autonomous operation. In this mode, sensors' data are used for building instantaneous maps that contain all parts of the environment observed by sensors, including dynamic obstacles. Figure 4 (b) exemplify an instantaneous occupancy grid map.

*C. Sensor Preprocessing*

Sensors capture data concurrently and at different rates. The first preprocessing step consists of synchronizing and grouping sensors' samples captured at nearly the same time. A queue is maintained for each sensor, except the less frequent one. When a sample from the less frequent sensor (camera if used, else Velodyne) is received, the most synchronized samples, i.e., the ones with smaller timestamp difference in relation to it, are returned from the queues and grouped together into a synchronized data package. Therefore, each data package contains a sample from odometry, IMU, GPS, Velodyne, and camera. A log is defined in this work as the set of data packages recorded while driving the self-driving car throughout an environment. The localization system estimates the car state using the most recent data package, while the mapping system receives as input the complete log.

After the synchronization and grouping step, the samples from each sensor are individually preprocessed. Biases in odometry measurements are compensated with the odometry calibration system presented by Mutz et al. (Mutz, et al., 2016). Calibration parameters are obtained by minimizing the average distance between GPS and dead-reckoning.

Velodyne point clouds are preprocessed for (i) eliminating invalid readings (measurements with range smaller than 1m or bigger than 70m), (ii) correcting points' positions considering the movement of the vehicle, and (iii) calibrating reflectivity measurements.

In order to transform Velodyne's points from the sensor coordinate system to the world coordinate system, the car pose has to be known. If the car is moving, its pose at the beginning of the sensor revolution may be different from the pose at the end of the revolution. To account for this fact, the car movement is estimated for each shot using the vehicle motion model (see Section IV.C) and this estimate is used to "correct" the pose before projecting the points into the map.

Due to physical properties (such as power and sensibility) of Velodyne's 32 lasers, they can return different reflectivity values when they hit objects with the same brightness. This lack of consistency among the lasers is problematic for both mapping and localization. A calibration similar to the one proposed by Levinson and Thrun (Levinson & Thrun, 2010) is used to correct reflectivity measurements. Our method receives as input a log and the car poses, and it outputs a calibration table whose cells store calibrated reflectivity values. We assume that the sensor calibration does not change over time, hence the table is built once using a log and then fixed.

The table is indexed by the laser identifier (0-31), the uncalibrated (measured) reflectivity (0-255), and a range-based index (0-9) obtained by discretizing the ray range in buckets that grow according to a geometric progression. In order to create the calibration table, a grid map is built using the log and the car poses. Each cell of the grid map stores all Velodyne measurements that hit the cell, along with their respective lasers' indices. We assume that all measurements that hit a given cells observed the same object, hence their reflectivity measurements should have been equal. This value is estimated using the average of the uncalibrated reflectivity measurements. These values are used to fill the calibration table. The assumption that rays that hit a cell observed the same object does not hold if the environment is dynamic since cells can be temporarily occupied by moving objects. Hence, it is preferable to perform the calibration in an environment without traffic.

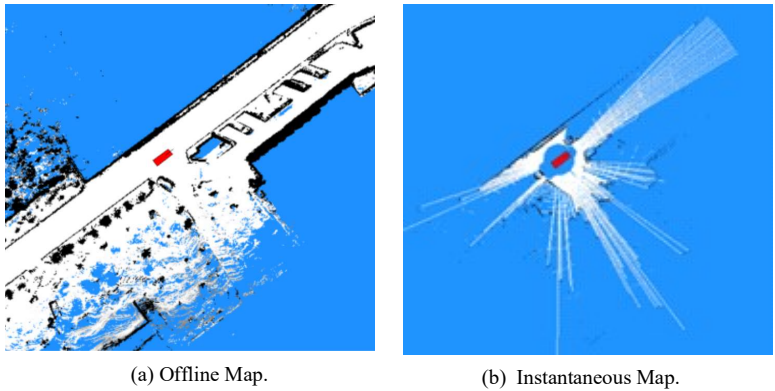

(a) Offline Map.   (b) Instantaneous Map.

Figure 4. Offline occupancy grid maps (a) and instantaneous occupancy grid maps (b). Offline maps are built before autonomous operation using a log of sensors' data. Instantaneous maps are built at runtime and used for estimating the localization of the self-driving car.

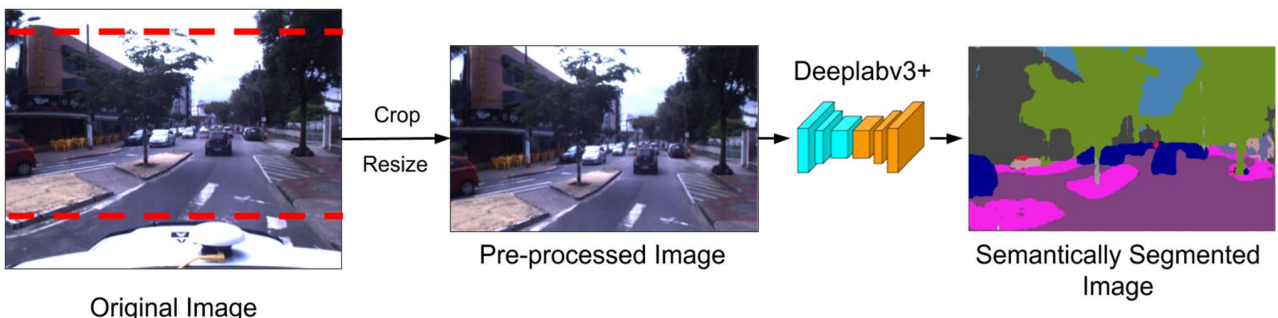

Figure 5. Preprocessing of images and visual semantic segmentation with DeepLabv3+ pre-trained in the Cityscapes dataset.

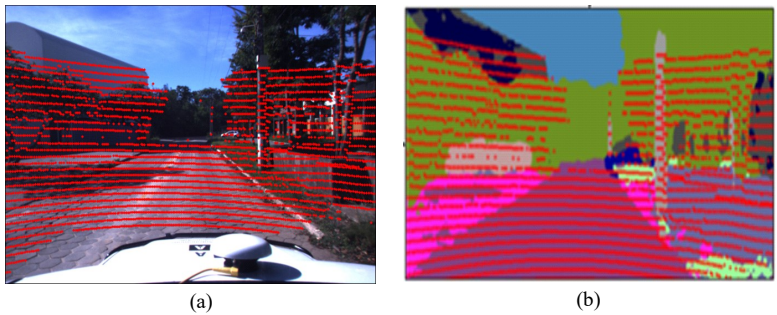

(a)   (b)

Figure 6. Projections a point cloud into an image captured by a front-facing camera (a) and its semantic segmentation (b). The points visible by the camera can be associated with the pixels' colors and/or labels. Note the slight misalignment between the point cloud and the image. Potential causes of this misalignement are desynchronization between sensors, latency in the camera, and imperfect extrinsic calibration.

Camera preprocessing consists of using a deep neural network, the DeepLabv3+ (Chen, Zhu, Papandreou, Schroff, & Adam, 2018) trained on the Cityscapes dataset (Cordts, et al., 2016) for extracting semantic information from the images. This process is illustrated in Figure 5. It was not necessary to fine-tune the model with our data. Before using the images as input to the neural network, they are cropped by 40 pixels in the top and 110 pixels in the bottom and resized to 513 x 264 pixels. This crop removes regions in the images that correspond in most part to the sky and the hood of the self-driving car.

Velodyne's point clouds are fused with images for obtaining a point cloud with precise range measurements, and colors and/or semantic labels. This sensor fusion is performed by projecting 3D points from Velodyne into image pixels (Hartley & Zisserman, 2003; Geiger, Lenz, Stiller, & Urtasun, 2013) and associating the point position with the pixel value. Figure 6 illustrates the projection of a point cloud into the raw image (Figure 6 (a)) and their respective semantic segmentation (Figure 6 (b)).

## IV. THE LOCALIZATION SYSTEM

This section presents the proposed localization system. It starts by reviewing the concept of grid maps and the type of information stored in occupancy, reflectivity, semantic, and color grid maps. It follows describing the particle filter, and its prediction and correction steps.

*A. Grid Maps*

In grid maps (Thrun, Burgard, & Fox, 2005; Thrun, 2002), the environment is discretized in cells with a fixed predefined size. Under the assumption that cars move in the ground plane and aiming at minimizing the use of computational resources, we employ 2D grid maps. Cells of a grid map usually store statistics that are updated as sensors observe the environment. Different statistics can be stored depending on the type of map.

Reflectivity grid maps (Badue, et al., 2020; Levinson & Thrun, 2010; Wolcott & Eustice, 2017) and colour grid maps (Bârsan, Liu, Pollefeys, & Geiger, 2018; Hornung, Wurm, Bennewitz, Stachniss, & Burgard, 2013) store, for each cell, the mean and variance of a Gaussian distribution (or the parameters of a Mixture of Gaussians (Wolcott & Eustice, 2017)) representing, respectively, the reflectivity or the colour of the objects inside the cell. Figure 1 (c) and Figure 1 (e) exemplify the reflectivity and color grid maps for the region presented in Figure 1 (a). The figures show the mean reflectivity and color and the variance is not represented.

In semantic grid maps (Kostavelis & Gasteratos, 2015; Yang, Huang, & Scherer, 2017; McCormac, Handa, Davison, & Leutenegger, 2017), cells are classified according to the types of objects they contain. Common classes of objects in self-driving car applications are road, building, tree, traffic sign, traffic light, and pedestrian. A map cell may receive several samples from different classes. In order to preserve all of this information, each cell stores a categorical distribution over classes of objects. Figure 1 (d) shows the semantic grid map for the region in Figure 1 (a). In this figure, the color of each cell represent the most likely class. Figure 7 illustrates the contents of a cell in a semantic grid map.

Occupancy grid maps (Thrun, Burgard, & Fox, 2005) store the probabilities of cells being occupied by obstacles. Occupancy grid maps are a special case of semantic grid maps with two classes, drivable (free) or not drivable (occupied). Figure 1 (b) shows the occupancy grid map for the region in Figure 1 (a). The grayscale color of the cells represent the occupancy probability with black being occupied and white being free cells.

Occupancy and reflectivity grid maps are built using data from the Light Detection and Ranging (LiDAR) sensor, while color and semantic grid maps use the point cloud obtained by fusing LiDAR and camera. Occupancy and reflectivity grid maps are traditionally employed for self-driving cars (Badue, et al., 2020) and they serve as a baseline in this work.

*B. Particle Filter*

The localization employs a particle filter (Thrun, Burgard, & Fox, 2005) for estimating the car state. In this work, the state $s = (v, \varphi, b, p)$ is composed of linear velocity $v$, steering wheel angle $\varphi$, steering wheel angle bias $b$, and car pose $p$. The car is assumed to move in the ground plane ($z = 0$), and the pitch and roll angles are assumed to be perfectly measured by the IMU. Given these assumptions, the pose reduces to a triplet $p = (x, y, \theta)$, in which $x$ and $y$ are the car position in the ground plane, and $\theta$ is the heading direction (yaw angle).

The particle filter approximates the probability distribution over the state using a set of samples (the particles). Each sample represents a possible value of the state. Weights are associated with the samples representing their likelihoods. Samples and weights are iteratively updated as new sensor data are observed.

Each update consists of a prediction step, followed by a correction step. In the prediction step, the values of the particles are adjusted to account for changes in the state (e.g., due to the movement of the car) since the last localization iteration. In the correction step, particles' likelihoods are evaluated and a new set of particles is generated by resampling (Thrun, Burgard, & Fox, 2005). A point estimate of the state can be obtained by computing the expected value of the particles.

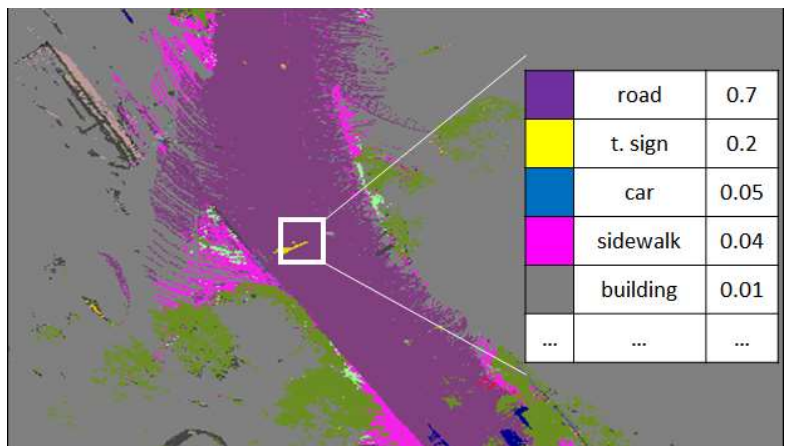

Figure 7. Illustration of semantic grid maps and the contents of their cells.

The particle filter is initialized as follows. We assume that an initial guess for the car pose is given either by GPS or by interacting with a user (e.g., the user manually informs the pose). Particles' poses are sampled from a Gaussian distribution with mean equals to the initial guess, and covariance given by the error of the guess (e.g., the uncertainty of GPS measurements informed by the manufacturer). The remaining components, linear velocity, steering wheel angle, and bias in steering wheel angle are sampled from Gaussian distributions with zero mean and predefined variances chosen by an expert.

## C. Prediction

In the prediction step, odometry data are used for updating the particles' states. Let $s_{t-1}^i = (v_{t-1}^i, b_{t-1}^i, \varphi_{t-1}^i, p_{t-1}^i)$ be the state of the $i$-th particle at time $t-1$. Then, the predicted state value, $\bar{s}_t^i$, given odometry data from time $t$, is given by:

$$\bar{s}_t^i = \begin{bmatrix} \bar{v}_t^i \\ \bar{b}_t^i \\ \bar{\varphi}_t^i \\ \bar{p}_t^i \end{bmatrix} = \begin{bmatrix} v_t \\ b_{t-1}^i \\ \varphi_t + \bar{b}_t^i \\ p_{t-1}^i \end{bmatrix} + \begin{bmatrix} \epsilon_v \\ \epsilon_b \\ \epsilon_\varphi \\ M_t^i + \epsilon_p \end{bmatrix} \quad (1)$$

where $v_t$ and $\varphi_t$ are the linear velocity and the steering wheel angle from the $t$-th data package; $\epsilon_v \sim \mathcal{N}(0, \sigma_v + \sigma_v^\varphi)$, $\epsilon_\varphi \sim \mathcal{N}(0, \sigma_\varphi + \sigma_\varphi^v)$, and $\epsilon_b \sim \mathcal{N}(0, \sigma_b)$, $\epsilon_p \sim \mathcal{N}(0, \Sigma_p)$ represent random noise obtained by sampling from Gaussian distributions. The value $\sigma_v$ is the uncertainty in $v_t$, $\sigma_v^\varphi$ is the uncertainty in $v_t$ due to $\varphi_t$, $\sigma_\varphi$ is the uncertainty in $\varphi_t$, $\sigma_\varphi^v$ is the uncertainty in $\varphi_t$ due to $v_t$, $\sigma_b$ is the uncertainty in $b_t$, and $\Sigma_p$ is the covariance matrix for the pose. $M_t^i$ is an estimate of the car movement since the last localization iteration obtained using odometry as input to the Ackermann motion model (Siegwart, Nourbakhsh, & Scaramuzza, 2011).

Assume that in time $t$, the car pose is $(x_t, y_t, \theta_t)$ and then the car moves for $\Delta_t$ seconds with linear velocity $v$ and steering wheel angle $\varphi$. According to the Ackermann motion model, the car pose after the movement $(x_{t+1}, y_{t+1}, \theta_{t+1})$ is given by:

$$x_{t+1} = x_t + v\,\Delta_t\,\cos(\theta_t) \quad (2)$$

$$y_{t+1} = y_t + v\,\Delta_t\,\sin(\theta_t) \quad (3)$$

$$\theta_{t+1} = \theta_t + v\,\Delta_t\,\frac{\tan(\varphi)}{L} \quad (4)$$

where $L$ is the distance between front and rear axles.

The term $\epsilon_p$ is added to account for physical phenomena not considered in the motion model (e.g., drift) and imprecisions in the map representation (e.g., due to discretization of the world in cells).

The update to the bias of the steering wheel angle is only applied when $\bar{v}_t^i$ is higher than a threshold (0.2m/s in this work). If this is not the case, the bias is maintained as it is. The value $\bar{b}_t^i$ is clipped to prevent it from growing to unrealistic values.

## D. Correction

The correction step is itself subdivided into likelihood computation and resampling. Likelihood computation consists of assigning weights to particles such that they receive high weights if they are likely to be the true state of the vehicle. Resampling is the process of generating a new set of particles from the previous ones considering the particles' weights. In this work, we employ low variance resampling (Thrun, Burgard, & Fox, 2005) for generating new particles.

In occupancy, semantic and reflectivity grid maps, the weight of the $i$-th particle, $s^i$, is given by the likelihood of observing point cloud $C$ assuming that the vehicle is in state $s^i$ and the offline map is $M$ (Figure 4 (a)). For increasing computational performance and to reduce the effect of noisy measurements, an instantaneous grid map $Z$ (Figure 4 (b)) is built using $C$ and used to approximate the likelihood of $C$ given $M$ and $s^i$:

$$p(C \mid s^i, M) \approx p(Z \mid s^i, M) \quad (5)$$

The probability density $p(Z \mid s^i, M)$ can be understood as follows. If a particle is correct, it will lead to an instantaneous map that match the offline map. On the other hand, if the particle is incorrect, its respective instantaneous map will be rotated or shifted in relation to the offline map. Therefore, by comparing the maps, we can indirectly evaluate the correctness of the particle.

Assuming that cells from the instantaneous map are independent, the conditional likelihood of $Z$ can be rewritten as the product of conditional likelihoods of cells:

$$p(Z \mid M, s^i) = \prod_{j=1}^{k} p(z_j \mid s^i, M) \quad (6)$$

where $k$ is the number of cells in $Z$. Since more than one ray can hit the same cell, $k \leq n$.

For preventing numerical issues and early convergence, the log function is used to transform the product into a sum of log likelihoods. Then, the weights of the $n$ particles, $w_{1:n}$, are obtained by normalizing their log likelihoods.

$$w_i = \frac{1}{\sum_{l=1}^{n} w_l} \sum_{j=1}^{k} \log p(z_j \mid s^i, M) \tag{7}$$

In order to obtain $p(z_j \mid s^i, M)$, we first compute the position of $z_j$ in the world coordinate frame using $s^i$. Let $z_j^i$ be the $j$-th cell after this transformation and $m_j^i$ be the cell in the same position in the offline map. The values of $z_j^i$ and $m_j^i$ are compared to evaluate the likelihood of observing $z_j^i$ given that the world is like $m_j^i$. This evaluation depends on the type of map used by the system.

In semantic grid maps and occupancy grid maps, the likelihood is obtained by selecting the class (free or occupied in occupancy grid maps; object types in semantic grid maps) with highest probability in $z_j^i$ and evaluating its probability in $m_j^i$. For instance, if most rays that hit $z_j^i$ are from the class "car", we return the "car" probability in $m_j^i$. Likewise, for occupancy grid maps, if $z_j^i$ is more likely to be occupied, then $p(z_j \mid s^i, M)$ is given by the occupancy probability of $m_j^i$.

For reflectivity grid maps, the Mahalanobis distance is employed. This is equivalent (up to a constant) to computing the log likelihood of a measurement in a Gaussian distribution. In our case, the measurement is the mean reflectivity of rays that hit $z_j^i$ and the Gaussian distribution is defined by $m_j^i$.

For color grid maps, a different approach is used. The particles weights are given by the normalized entropy correlation coefficient (ECC) (Astola & Virtanen, 1983; Thanh, Kim, Hong, & Ngo Lam, 2018) between the instantaneous and offline maps. The notation $x_{a:b}$ is used from now on to represent the sequence $\{x_a, \dots, x_b\}, a \leq b$. Let $z_{1:k}^i$ be the observed cells in $Z$ in the world coordinate system and $m_{1:k}^i$ be the cells in the same rows and columns in $M$. Then, the weight of the $i$-th particle is given by:

$$w_i = \frac{1}{\sum_{l=1}^{n} w_l} ECC(z_{1:k}^i, m_{1:k}^i) \tag{8}$$

ECC is chosen instead of Mahalanobis distance to minimize negative effects of different illumination conditions. The value of a cell can change in color grid maps if there are shadows over the cell or if the offline map was built in daylight and the car is operating at night, for example. ECC is defined by:

$$ECC(z_{1:k}^i, m_{1:k}^i) = 2 - \frac{2\,H(z_{1:k}^i, m_{1:k}^i)}{H(z_{1:k}^i) + H(m_{1:k}^i)} \tag{9}$$

where $H(z_{1:k}^i)$ and $H(m_{1:k}^i)$ are the marginal entropies of $z_{1:k}^i$ and $m_{1:k}^i$, and $H(z_{1:k}^i, m_{1:k}^i)$ is their joint entropy.

The metric is scaled to $(0, 1)$, with 0 indicating full independence and 1 representing complete dependence between variables. ECC is related to the normalized mutual information and both were vastly employed for registration of images from different sources (e.g., medical images from different sensors) (Amaral-Silva, Murta-Jr, de Azevedo-Marques, Wichert-Ana, & Studholme, 2016; Bardera, Feixas, Boada, & Sbert, 2006; Cahill, 2010).

The entropy terms can be computed using histograms of pixels intensities in grayscale as described in (Veronese, et al., 2016). Let $H^z$ be the histogram of the grayscale $z_{1:k}^i$ with $n$ bins and $h_l^z$ be the probability of the $l$-th bin. Then, $H(z_{1:k}^i)$ is given by [1]:

$$H(z_{1:k}^i) = -\sum_{l=1}^{n} h_l^z \log(h_l^z) \tag{10}$$

$H(m_{1:k}^i)$ can be computed analogously from the grayscale version of $m_{1:k}^i$. The joint entropy $H(z_{1:k}^i, m_{1:k}^i)$ is obtained by building a joint histogram $H^{z,m}$. The joint histogram is a matrix in which rows and columns refer to intervals of grayscale values in the offline and instantaneous map, respectively. Each cell in the matrix stores the number of times two grayscale values were observed "together" in the same map cell. Assume that $H^{z,m}$ has $p$ rows and $q$ column and that $h_{r,s}^{z,m}$ is the normalized frequency (the cell count divided by the sum of all values from the matrix) of the cell in row $r$ and column $s$. Then, $H(z_{1:k}^i, m_{1:k}^i)$ is given by:

$$H(z_{1:k}^i, m_{1:k}^i) = -\sum_{r=1}^{p} \sum_{s=1}^{q} h_{r,s}^{z,m} \log(h_{r,s}^{z,m}) \tag{11}$$

---

[1] Note that equations 10 and 11 are special cases of the general definition of entropy (Thrun, Burgard, & Fox, Probabilistic robotics, 2005).

Rays that hit moving obstacles or structures that were not present during mapping can lead to incoherent particles' likelihoods. Due to noisy measurements (outliers), incorrect particles may receive high weights, while low weights may be assigned to correct particles. To minimize this issue, an outlier rejection approach is employed. If the likelihood of a sensor measurement is lower than a threshold for most particles, then the measurement is not considered when computing the particles' likelihoods.

## V. THE MAPPING SYSTEM

A block diagram representing the mapping system is shown in Figure 8. It receives as input a log of sensors' data and returns as output reflectivity, occupancy, color, and semantic grid maps.

After performing the steps described in Section III.C, additional preprocessing is applied to the log when mapping. Data packages are discarded if the car is stopped, i.e., if its speed is smaller than a threshold, or if the Euclidean distance between consecutive GPS measurements is larger than a threshold (under the assumption that a jump has happened). The remaining data packages are used as input for the pose estimation process that outputs the poses of the self-driving car for each data package. Finally, point clouds are projected into grid maps using the poses.

Since *a priori* maps are not available at this point, poses have to be estimated using Simultaneous Localization and Mapping (SLAM) (Thrun, Burgard, & Fox, 2005). Given that pose estimation can be performed offline before autonomous operation, we employ full SLAM techniques that use all information available in the log for obtaining accurate estimates. Previous full SLAM algorithms (Levinson, Montemerlo, & Thrun, 2007; Levinson & Thrun, 2010; Wolcott & Eustice, 2017; Mutz, et al., 2016) lead to maps with imprecise loop closures in our experiments. Therefore, we developed a novel two-step process based on GraphSLAM (Thrun & Montemerlo, 2006) for estimating the car path.

The pose estimation process is composed of a sequence of steps. A block diagram representing it is presented in Figure 9. It receives as input data from GPS, odometry, and Velodyne, and returns as output car poses. Since IMU is only used for measuring pitch and roll angles, it is not depicted in the figure.

First, data from odometry and GPS are fused for producing an initial guess for pose values. This initial guess is called fused odometry. It is used for detecting loop closure regions and, along with Velodyne data, for estimating the map misalignment in these regions.

Final values of the poses are obtained using GPS, odometry, and loop closure constraints as input for GraphSLAM. In GraphSLAM, nodes represent the variables to be estimated and edges represent observations. In this work, nodes are vehicle poses, and edges are measurements from sensors and their covariances (assuming that measurements follow Gaussian distributions).

We employ two types of edges, unary and binary edges. Unary edges represent global pose measurements in the world coordinate system, such as those from GPS. They define constraints of the form:

$$G(x, z) = (x - z)^T C^{-1} (x - z) \quad (12)$$

where $x$ is the pose being estimated, $z$ is a pose measurement and $C$ is the covariance matrix representing the uncertainty in $z$.

Binary local edges represent relative translation and rotation between two poses. Examples of measurements that lead to binary edges are loop closures and the movements of the car between two instants obtained using odometry data as input to the motion model. Binary edges imply in constraints of the form:

$$L(x_a, x_b, z) = (x_a - x_b - z)^T C^{-1} (x_a - x_b - z) \quad (13)$$

where $x_a$ and $x_b$ are poses being estimated, $z$ is the relative transformation between $x_a$ and $x_b$, $C$ is the covariance matrix for $z$.

Measurement $z$ in equation 13 can be interpreted as the difference (in poses; not to be confused with real difference) between $x_a$ and $x_b$. Another interpretation is that $z$ is the value of $x_a$ in the coordinate system defined by $x_b$. Furthermore, if $x_a$, $x_b$ and $z$ are represented as 4x4 transformation matrices, we have that $x_a = x_b\ z$. In equations 12, 13, 15 17, and 19, plus sign represents the pose composition operator and minus sign represents its inverse (Corke, 2017).

In equation 12, doing $x = z$ minimizes $G$. Similarly, the values of $x_a$ and $x_b$ that minimize $L$ in equation 13 are those that satisfy $x_a = x_b + z$. In GraphSLAM, the objective function is defined by a set of constraints of these forms. Therefore, poses values resultant from the optimization trade-off local and global constraints. Covariance matrices control the relative importance or the "weight" of the constraints. They can also be interpreted as a soft constraint of the search space. If $z$ is highly accurate in equations 12, for instance, then the value of $x$ should not diverge much from it. Likewise, if $z$ is uncertain, the true value of $x$ may be far from $z$, hence the optimizer can move $x$ away from $z$ if there are other measurements with higher precision.

Next sections describe how these constraints are used in the two-step pose estimation process. In them, assume that the log used for mapping contains $n$ data packages, $p_{1:n}$, after preprocessing. The $i$-th data package is $p_i = \{g_i, o_i, c_i, v_i\}$, where $g_i, o_i, c_i, v_i$ correspond to data from GPS, odometry, camera, and Velodyne, respectively. The car's poses when the data packages were acquired will be referred to as $x_1, \dots, x_n$.

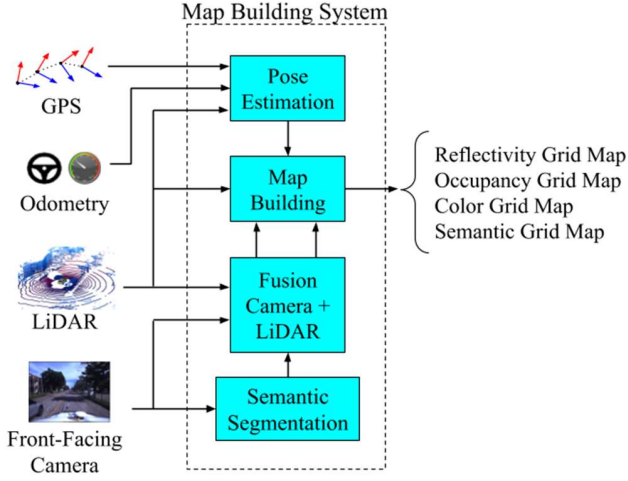
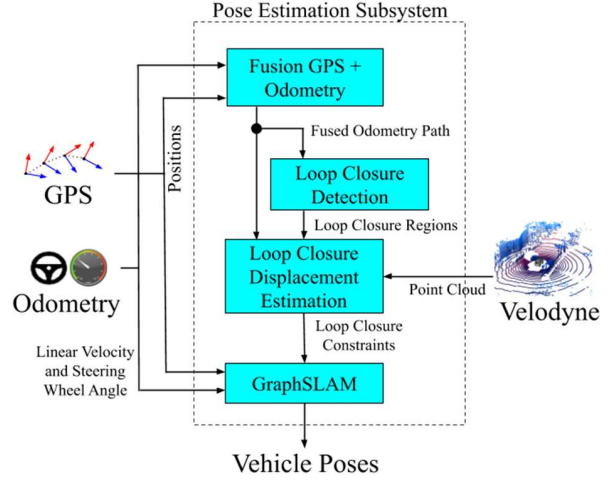

Figure 8. Block diagram describing the mapping system.    Figure 9. Block diagram describing the pose estimation process.

*A. Fused Odometry*

In first step, pose estimates $\hat{x}_{1:n}$ are computed fusing GPS and odometry. Data from GPS enforces global consistency while odometry and the motion model encourages smooth transitions between poses. The values of $\hat{x}_{1:n}$ are obtained by minimizing the objective function $F$ which is given by:

$$F = \sum_{t=1}^{n} G(x_t, g_t) + \sum_{t=2}^{n} L(x_{t-1}, x_t, \delta_t) \tag{14}$$

where $g_t$ is the GPS pose at time $t$, and $\delta_t$ is the motion estimate obtained using the odometry measurement as input to the Ackermann motion model.

Expanding the terms of the sum we have:

$$F = \sum_{t=1}^{n} (x_t - g_t)^T R_t^{-1} (x_t - g_t) \\ + \sum_{t=2}^{n} (x_t - x_{t-1} - \delta_t)^T Q_t^{-1} (x_t - x_{t-1} - \delta_t) \tag{15}$$

where $R_t$ and $Q_t$ are covariance matrices representing the uncertainties in $g_t$ and $\delta_t$, respectively.

*B. Loop Closures Detection and Displacement Estimation*

Due to imprecisions in GPS and odometry measurements, poses' values obtained in the first step of optimization can be imperfect. Because of that, if the self-driving car moves over the same region more than once, projections of sensor data from consecutive visits into the grid map may not match. This inconsistency results in a "ghosting effect" observed as drifts, duplication, or blurring of objects in the map. This issue is known as the loop closure problem (Badue, et al., 2020; Thrun, Burgard, & Fox, 2005) and illustrated in Figure 10 (a).

To account for this problem, we run the second step of pose estimation using GraphSLAM. Like the previous step, GPS and odometry data are used to constrain the pose values globally and locally, respectively. However, besides them, loop closure constraints are also considered to penalize poses' values whose sensor projections into the map do not match. Figure 10 (b) shows the map built considering loop closures in the pose estimation. Note that artefacts visible in Figure 10 (a) are no longer present. In order to add loop closure constraints into the graph, loop closures must be detected and their displacements must be estimated.

Previous works attempted to use GPS for detecting loop closures (Mutz, et al., 2016). However, correct loop closures may be missed and poses that are distant from each other may be incorrectly classified as loop closures due to errors and jumps in GPS. Reducing these errors and jumps is the main motivation for the first step of optimization.

Our loop closure detection technique uses the fused odometry and it is simple yet effective. Each pose $\hat{x}_t$ is classified as being part of a loop closure or not. To do so, we search if there are poses in $\hat{x}_{1:t-1}$ (the set of previous poses) such that their Euclidean distances to $\hat{x}_t$ are smaller than a threshold $\tau_d$ and their distance in time (difference of timestamps) to $\hat{x}_t$ are larger than a threshold $\tau_t$. If there are poses that satisfy both conditions, $\hat{x}_t$ is classified as being part of a loop closure. The second condition prevent the classification of consecutive poses as loop closures since they are close to each other. If there are no poses that meet both conditions, we assume the self-driving car is visiting the region for the first time.

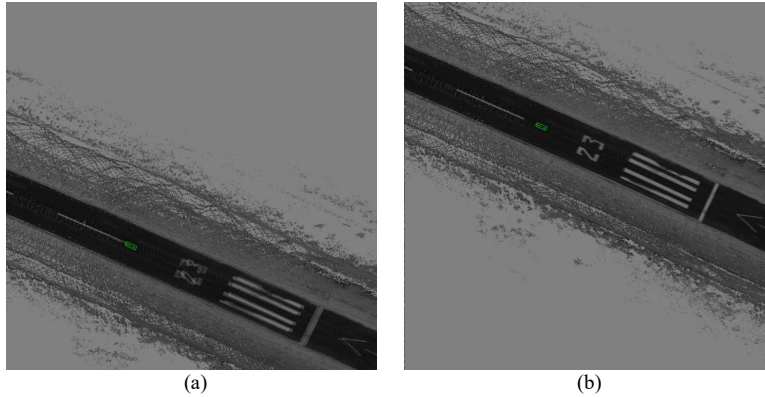

(a)      (b)

Figure 10. (a) Ghosting effect observed when revisiting a region if fused odometry poses are used for projecting sensor data into the grid map. (b) The same region after considering loop closures in the state estimation process.

Poses classified as being the first visit to a region and their associated data packages are used for building reflectivity and occupancy grid maps. Edges are added into the graph connecting every pose classified as a loop closure and its nearest pose. The displacement between the poses is computed by using the localization technique (Section IV) for estimating the car pose in relation to the map of the first visit.

Since the localization technique can fail, we try and filter out outliers by comparing localization poses and their respective fused odometry poses. If their Euclidean distance or their difference in orientation are higher than predefined thresholds, the loop closure constraints are discarded.

Semantic and colour grid maps were also considered in the process of handling loop closures in preliminary experiments. However, they were not accurate enough for improving the displacement estimation, and adding their constraints into the graph caused divergence.

The final values of the poses, $x^*_{1:n}$, are obtained by minimizing the objective function $J$ which is given by:

$$J = F + \sum_{(x_t, x_s) \in LC} L(x_t, x_s, \tilde{x}^s_t) \quad (16)$$

where $F$ is the objective function of the first step of optimization (Equation 15), $LC = \{(x_t, x_s)\}$ is a set in which each $x_t$ correspond to a pose classified as a loop closure, $x_s$ is its nearest pose in $\hat{x}_{1:t-1}$, and $\tilde{x}^s_t$ is the relative pose between them estimated with the localization.

Expanding the terms, we have:

$$J = \sum_{t=1}^{n} (x_t - g_t)^T R_t^{-1} (x_t - g_t)$$
$$+ \sum_{t=2}^{n} (x_t - x_{t-1} - \delta_t)^T Q_t^{-1} (x_t - x_{t-1} - \delta_t) \quad (17)$$
$$+ \sum_{(x_t, x_s) \in LC} (x_t - x_s - \tilde{x}^s_t)^T S_t^{-1} (x_t - x_s - \tilde{x}^s_t)$$

where $S_t$ the covariance matrix of $\tilde{x}^s_t$.

### C. Map Building

The vehicle poses obtained by the previous optimization process are used to project LiDAR point clouds (either raw or fused with images) to the world (global) coordinate system and then into cells of the grid maps. Points that hit the same cell are used to compute cell statistics.

For reflectivity and color grid maps, mean and standard deviation are estimated. In semantic grid maps, each cell stores a categorical distribution over classes of objects. This distribution is represented as a vector in which the $i$-th dimension represents the probability that objects of class $i$ exist in the cell. Instead of storing actual probabilities, in our implementation, we store counters representing the number of times each class was observed in the cell. Probabilities are derived from counters by normalization when necessary. The advantage of storing counters instead of actual probabilities is that cells can be updated by simple incrementing counters.

To initialize semantic grid maps, we assume a uniform prior distribution, and the counters of all cells are set to one. Although initializing the counters with zero seems more intuitive since cells were not observed by sensors yet, it is problematic for two reasons. First, it would lead to division by zero when normalizing the cells. Second, since particles' weights in localization are computed as a product of likelihoods, the initialization of counters with zero can make particles receive zero weight because a single point of the point cloud has zero likelihood.

In occupancy grid maps, two updates are performed on the map for each range measurement. In the first update, the occupancy probability of the cell hit by the laser is updated according to the evidence that the cell contains an obstacle (computed as in (Mutz, et al., 2016)). In the second update, a raycast technique (Wolcott & Eustice, 2017; Thrun, Burgard, & Fox, 2005; Mutz, et al., 2016; Badue, et al., 2020) decreases the occupancy probabilities of all cells between the first ray of a shot that hit the floor and the first ray that hit an obstacle. This update is based on the fact that if obstacles were present in any of these cells, a smaller range would have been measured. Raycast makes occupancy grid maps denser than other types of maps since several cells are updated for each point.

For limiting the amount of memory used by the system, grid maps are stored and used as a set of squared fixed-sized tiles. Only the tile that currently contains the car along with the eight tiles around it are maintained in primary memory (for a more detailed description, refer to (Veronese, et al., 2016)).

## VI. EXPERIMENTAL EVALUATION

This section presents datasets and the experimental methodology employed for evaluating the mapping and localization techniques. Results achieved are also presented and discussed.

### A. Datasets

Four large-scale environments are considered in experiments. For each environment, a dataset is built. In this work, a dataset is defined as a set of logs, one of which is used for mapping, and the remaining for testing the localization. Logs are named using an environment identifier followed by a suffix that represents their purpose and main features. Logs with suffix "M" are used for mapping. These logs are usually recorded in good illumination conditions, low traffic and few parked vehicles. Logs with suffix "TM" are test logs built in the same conditions as those used for mapping. Suffix "TD" indicates test logs with hundreds of distractors (e.g. parked and moving vehicles, pedestrians, and environmental changes). Finally, test logs with the suffix "TN" were recorded at night (after 7 p.m.). The paths of the vehicle in the datasets are presented in Figure 11. The datasets are described next.

*1) UFES:* logs recorded in the ring road of the Federal University of Espírito Santo (Universidade Federal do Espírito Santo – UFES). It consists of a mapping log, UFES_M, and two test logs, UFES_TD and UFES_TN. Both test logs were recorded months after the mapping log. Figure 12 illustrates different illumination conditions present in the logs and their semantic segmentations. Note the different number of parked cars in the logs, and the presence of light blobs and motion blur at night.

*2) AIRPRT:* dataset built in the runway of an airport. Because the access to the airport is restricted, the mapping and test logs, AIRPRT_M and AIRPRT_TM, respectively, were created on the same day at sunrise (about 5 a.m.). This dataset allows evaluating the capacity of the localization technique of operating in adverse conditions of illumination (not as harsh as the logs recorded at night though). Besides that, some highways are similar to an airport runway in the sense that they consist of a flat road with eventual lane marks and with a surrounding area with few features (usually vegetation) for localization.

*3) PARK*: logs recorded in a parking lot at UFES. The parking lot is not part of the ring road. The dataset comprises a mapping log, PARK_M, and two test logs, PARK_TD and PARK_TN. The mapping log was built with an empty parking lot. In both test logs, there were tens of parked cars. This dataset allows evaluating the robustness of the localization technique in face of tight U-turns and with objects that were not present during mapping.

*4) URBAN:* dataset built in the neighborhood of Jardim da Penha, in the City of Vitória, ES, Brazil. The dataset contains a mapping log, URBAN_M, and two test logs, URBAN_TD and URBAN_TN. The environment contains diverse urban scenarios such as tree-lined streets, tunnels, residential areas, loops around plazas, and an avenue with intense traffic.

### B. Localization Ground Truth

Producing a ground truth for evaluating the localization is quite challenging. Although globally consistent, GPS data can be subject to significant amounts of noise, and GPS measurements are not necessarily consistent with the map. Maps are built using poses that result from an optimization process. These poses can drift locally from GPS which can make maps inconsistent with the real world. Since the localization is computed in relation to the map, the localization estimate may disagree with the GPS, while being correct with respect to the map. Therefore, comparing localization estimates with GPS data leads to a poor evaluation. Likewise, due to noise in GPS and in odometry measurements, paths estimated (using the methods presented in Section V.A and section V.B) from two logs of the same region can be locally inconsistent with each other. Hence, poses from the test log obtained by the optimization process cannot be used as ground truth. Since the method does not consider the map, there are no guarantees that the poses will match the localization in relation to the map.

The approach used in our work is building the ground truth by considering the map into the optimization. Additional edges are added to the graph to encourage test poses to be consistent with the map even if the map is locally imperfect. This approach is similar to ones found in previous works (Levinson & Thrun, 2010; Wolcott & Eustice, 2017; Levinson, Montemerlo, & Thrun, 2007).

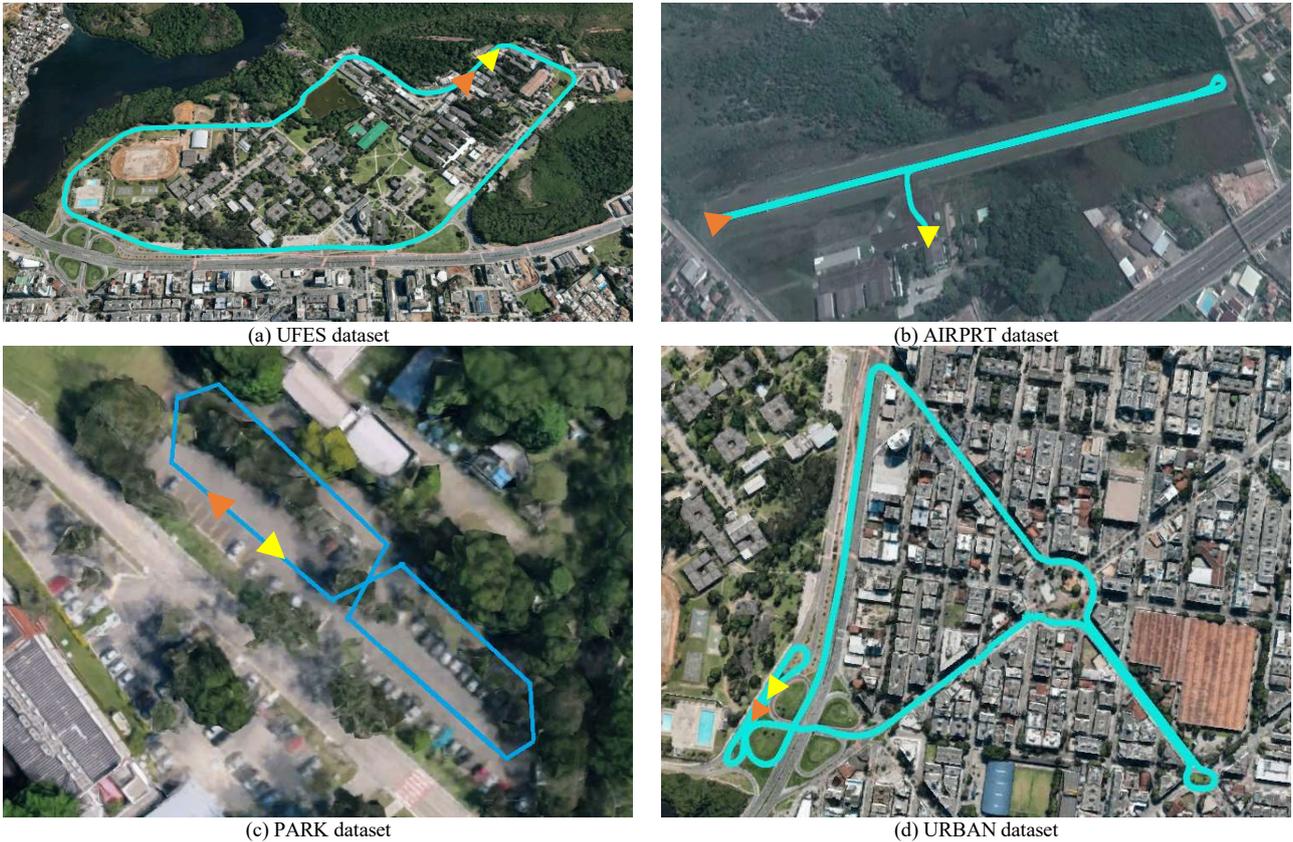

Figure 11. Satellite view of environments used in experiments. Starting poses are represented by orange triangles, while end poses are drawn in yellow.

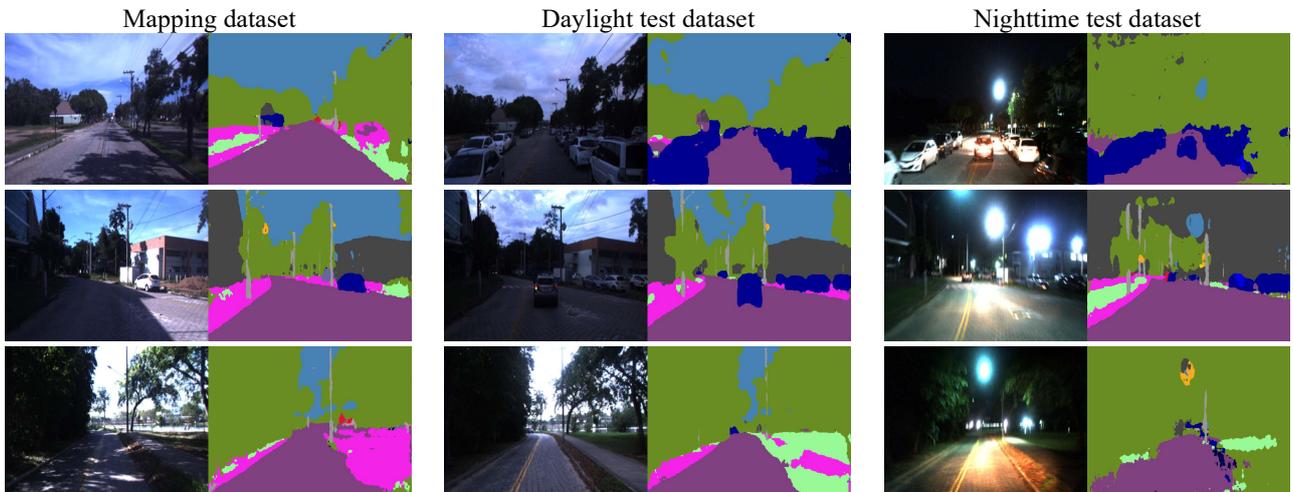

Figure 12. Challenging illumination conditions present in the UFES dataset. Each column corresponds to the same location. Even at night and with shadows, the deep neural network DeepLabv3+ managed to produce reasonable semantic segmentations

Consider two logs, one for mapping and one for testing the localization. Poses of the mapping log are obtained using the technique presented in Section V and used for building grid maps. From this point onwards, these poses are fixed. Ground truth poses for the test log are also obtained by optimization, but adding global edges into the graph to encourage consistency with the maps. Measurements of these edges are not collected from sensors. They are obtained by estimating the localization in reflectivity and occupancy grid maps. Note as this approach is similar to the one used for handling loop closures. A key difference, however, is that mapping poses are fixed. This would be equivalent to fixing the poses of the first visit to a region and only optimizing the poses from the second visit when handling loop closures.

The final graph for obtaining the ground truth poses considers data from GPS, odometry, loop closures and the new global edges obtained with localization. By using the sensors' data together with the localization estimates, we can compensate eventual localization errors. This graph induces the following objective function (using the same notation as in Section V):

$$Q = J + \sum_{t=1}^{n} G(x_t, l_t) \qquad (18)$$

where $J$ is the objective function from equation 17, $l_{1:n}$ are the localization estimates in relation to the maps, and $G$ is the function defined in equation 12.

Expanding the terms of the sum, we have:

$$\begin{aligned}
Q = &\sum_{t=1}^{n} (x_t - g_t)^T R_t^{-1} (x_t - g_t) \\
&+ \sum_{t=2}^{n} (x_t - x_{t-1} - \delta_t)^T Q_t^{-1} (x_t - x_{t-1} - \delta_t) \\
&+ \sum_{(x_t, x_S) \in LC} (x_t - x_S - \tilde{x}_t^S)^T S_t^{-1} (x_t - x_S - \tilde{x}_t^S) \\
&+ \sum_{t=1}^{n} (x_t - l_t)^T L_t^{-1} (x_t - l_t)
\end{aligned} \qquad (19)$$

where $L_t$ is the covariance matrix associated with $l_t$.

The poses values that result from optimizing $Q$ are the taken to be the localization ground truth.

*C. Experimental Methodology*

Experiments comprise qualitative analyses of maps built using the mapping technique and quantitative analyses of the localization technique in diverse conditions of operation. The metrics for evaluating the localization are the root mean squared error (RMSE) and the mean standard deviation (STD) to the ground truth. Bresson et al. argue that for operating safely, localization errors should never be bigger than 20cm (Bresson, Alsayed, Yu, & Glaser, 2017). Inspired by this observation, we also present for each test log the percentage of data packages in which the localization error is smaller than the thresholds 0.5m, 1.0m, and 2.0m.

Table 1 presents the values of parameters used in the experiments. Localization parameters were manually adjusted using a log from UFES different from the ones used for testing the system. Once chosen, the values were fixed for all experiments.

Two sets of parameters for position STD and orientation STD in the prediction step of localization are evaluated. The STABLE configuration favors stability or smoothness in the particle filter and the DIVERSE configuration favors recoverability or diversity. The different behaviors of the particle filter are chosen by using different amounts of noise to update the particles values during the prediction step. In the STABLE configuration, position STD and orientation STD are set to 0.01m and 0.1degrees. In the DIVERSE configuration, they are set to 0.2m and 0.5 degrees.

By adding more noise into particles' poses, the spreading of particles is increased which can increase the capacity of the filter of recovering from incorrect estimates. On the other hand, high levels of noise can increase the variance of pose estimates. The high variance can cause oscillations and abrupt movements during autonomous operation as a result of the decision-making system trying to adjust the vehicle trajectory given the new pose estimate.

*D. Results*

Grid maps built with the logs UFES_M, AIRPRT_M, PARK_M, and URBAN_M are presented in Figure 13. The proposed mapping technique was successfully employed for building all maps with consistent loop closures.

The quantitative evaluation of the localization technique is presented in Table 2. The localization based only on LiDAR (occupancy or reflectivity grid maps) led to smaller RMSE than configurations that require fusing LiDAR and camera. Semantic grid maps achieved, in general, errors close but slightly worse than occupancy or reflectivity grid maps. In the URBAN_TD dataset with STABLE configuration, using semantic grid maps was better than using occupancy grid maps (which diverged). In the same dataset, but with the DIVERSE configuration, semantic grid maps led to higher accuracy than reflectivity grid maps. The localization with color grid maps achieved a significantly lower accuracy than the other configurations even with a robust metric being used for computing the particles' weights. The non-uniformity of the illumination and the color variation due to it may have caused the negative results.

Table 1. Parameters for mapping and localization

| Parameters | Value |
|---|---|
| Map Resolution | 0.2m |
| Tile Size | 70m |
| STD for first pose (x, y) | 2.5m |
| STD for first pose (orientation) | 20deg |
| STD for reflectivity and color measurements | 3 |
| Number of particles | 200 |
| STD for linear velocity during prediction | 0.2m/s |
| STD for steering wheel angle during prediction | 0.5deg |
| Outlier rejection rate | 0.7 |

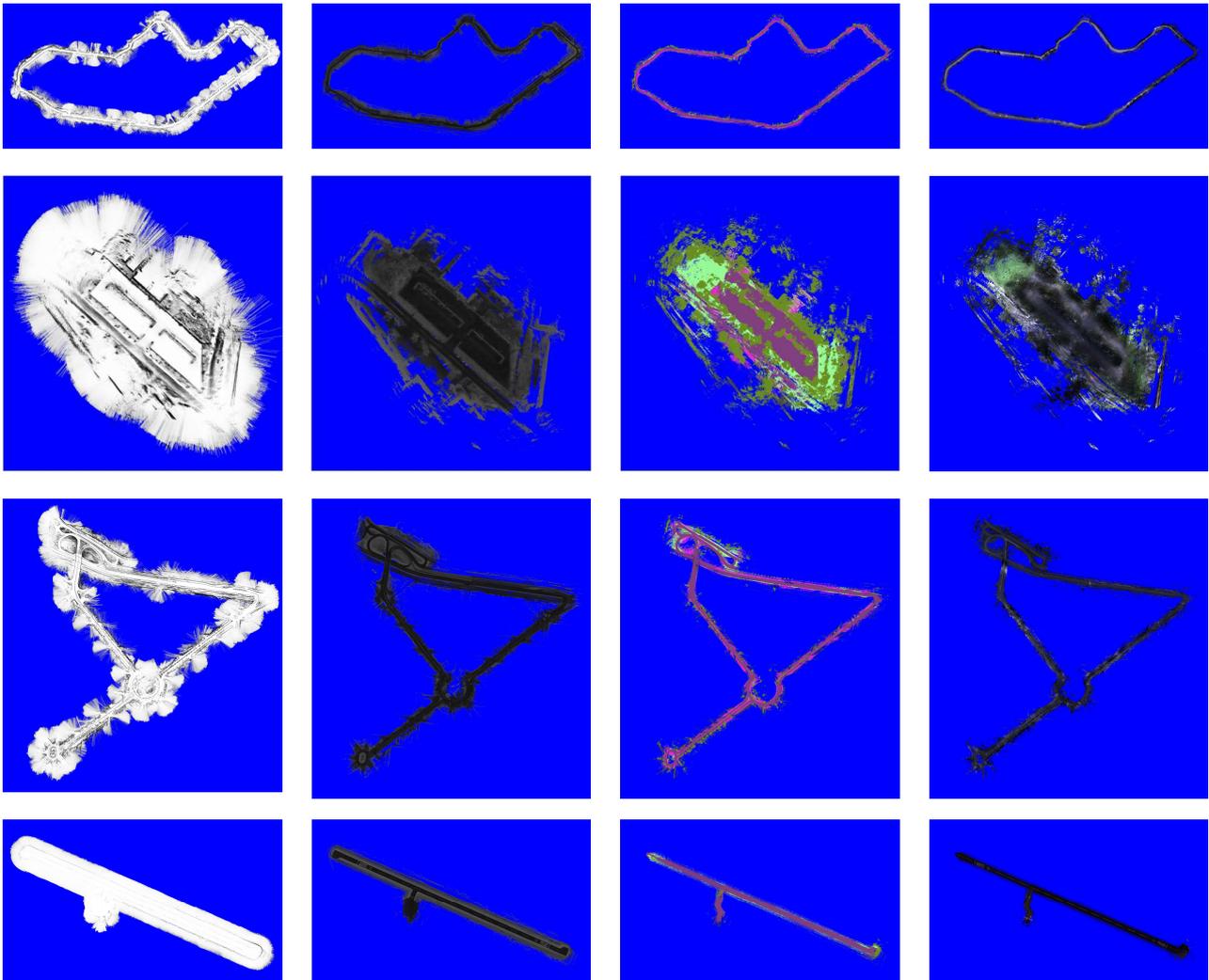

Figure 13. Grid maps built using the novel mapping technique. Rows represent the datasets UFES, PARK, URBAN, and AIRPRT, respectively, and from left to right, columns represent occupancy, reflectivity, semantic and colour grid maps.

Table 2. Localization accuracy when using different types of grid maps.

|  | RMSE (m) | STD (m) | % < 2m | % < 1m | % < 0.5m | RMSE (m) | STD (m) | % < 2m | % < 1m | % < 0.5m |
|---|---|---|---|---|---|---|---|---|---|---|
|  | DIVERSE | | | | | STABLE | | | | |
|  | UFES_TD | | | | | | | | | |
| Occupancy | **0.18** | **0.02** | **100.00** | **100.00** | **94.58** | 0.59 | 0.23 | 97.16 | 92.35 | 80.10 |
| Reflectivity | 0.19 | 0.02 | 100.00 | 100.00 | 93.83 | 0.55 | 0.19 | 97.85 | 93.05 | 86.18 |
| Semantic | 0.30 | 0.02 | 100.00 | 99.49 | 92.95 | 0.61 | 0.18 | 98.86 | 89.10 | 71.67 |
| Color | 1.85 | 2.64 | 89.03 | 83.39 | 68.27 | 11.28 | 44.17 | 18.77 | 16.14 | 14.17 |
|  | UFES_TN | | | | | | | | | |
| Occupancy | **0.19** | **0.02** | **100.00** | 99.98 | 96.17 | 0.28 | 0.05 | 100.00 | 97.74 | 93.90 |
| Reflectivity | 0.20 | 0.02 | 99.98 | 99.93 | 96.85 | 0.34 | 0.08 | 99.96 | 95.99 | 90.48 |
| Semantic | 0.51 | 0.10 | 99.44 | 95.20 | 76.73 | 1.86 | 2.54 | 85.13 | 80.48 | 70.87 |
| Color | 6.05 | 16.75 | 40.30 | 23.27 | 11.40 | 6.71 | 19.69 | 28.24 | 17.10 | 8.45 |
|  | PARK_TD | | | | | | | | | |
| Occupancy | 0.19 | 0.02 | 100.00 | 100.00 | 95.93 | 0.22 | 0.02 | 100.00 | 100.00 | 93.81 |
| Reflectivity | 0.17 | 0.01 | 100.00 | 100.00 | 98.99 | **0.17** | **0.01** | **100.00** | **100.00** | **99.33** |
| Semantic | 0.28 | 0.03 | 100.00 | 100.00 | 88.12 | 0.35 | 0.05 | 100.00 | 98.50 | 84.52 |
| Color | 20.94 | 216.08 | 22.62 | 18.54 | 12.60 | 9.61 | 9.11 | 0.54 | 0.30 | 0.24 |
|  | PARK_TN | | | | | | | | | |
| Occupancy | 0.22 | 0.02 | 100.00 | 100.00 | 95.68 | **0.20** | **0.02** | **100.00** | **100.00** | **96.15** |
| Reflectivity | 0.20 | 0.02 | 100.00 | 100.00 | 95.15 | 0.22 | 0.02 | 100.00 | 100.00 | 95.99 |
| Semantic | 0.63 | 0.17 | 98.32 | 93.34 | 63.48 | 0.34 | 0.03 | 100.00 | 100.00 | 88.44 |
| Color | 22.85 | 221.43 | 11.48 | 3.96 | 1.02 | 10.85 | 21.84 | 2.94 | 0.67 | 0.04 |
|  | URBAN_TD | | | | | | | | | |
| Occupancy | 83.49 | 5849.75 | 68.31 | 67.84 | 61.07 | **0.50** | **0.18** | **97.86** | **95.03** | **88.04** |
| Reflectivity | 0.52 | 0.18 | 98.34 | 93.72 | 82.32 | 0.83 | 0.52 | 95.55 | 90.77 | 81.96 |
| Semantic | 0.60 | 0.19 | 97.61 | 93.28 | 78.00 | 0.55 | 0.17 | 97.78 | 94.76 | 81.89 |
| Color | 17.46 | 246.27 | 60.80 | 59.71 | 55.08 | 0.60 | 0.20 | 97.88 | 92.16 | 78.48 |
|  | URBAN_TN | | | | | | | | | |
| Occupancy | **0.19** | **0.02** | **100.00** | **99.82** | **96.59** | 1.18 | 1.16 | 93.21 | 89.77 | 83.51 |
| Reflectivity | 0.22 | 0.03 | 100.00 | 98.12 | 96.51 | 2.10 | 2.75 | 69.66 | 63.87 | 59.99 |
| Semantic | 0.60 | 0.14 | 98.54 | 93.00 | 69.22 | 2.28 | 3.92 | 84.68 | 82.66 | 68.07 |
| Color | 11.79 | 57.33 | 14.14 | 8.46 | 4.61 | 29.44 | 297.68 | 0.78 | 0.08 | 0.05 |
|  | AIRPRT_TM | | | | | | | | | |
| Occupancy | 0.10 | 0.01 | 100.00 | 99.95 | 99.65 | **0.10** | **0.00** | **100.00** | **100.00** | **99.65** |
| Reflectivity | 0.25 | 0.05 | 100.00 | 97.89 | 97.19 | 0.10 | 0.00 | 100.00 | 99.85 | 99.75 |
| Semantic | 3.14 | 4.49 | 56.78 | 38.44 | 20.09 | 4.66 | 3.95 | 20.44 | 14.84 | 10.40 |
| Color | 7.37 | 3.99 | 0.82 | 0.35 | 0.12 | 7.16 | 4.00 | 0.94 | 0.23 | 0.12 |

The average errors were bigger with the STABLE configuration than with the DIVERSE configuration. This result supports the hypothesis that adding more noise during the prediction step increases the capacity of the method of recovering from momentary drifts. However, it is important to note that the best results in most datasets were achieved with the STABLE configuration.

The fact that the localization with semantic grid maps was more accurate than the localization with color grid maps is worth commenting since both rely on the same data (images and LiDAR). The most likely cause of the different performances is the neural network's capacity of performing coherent semantic segmentation even in face of harsh illumination conditions (e.g., at night or with shadows). The neural network managed to produce segmentations that were sufficiently good for estimating the self-driving car localization. Therefore, results show that given a robust method for extracting semantic information from sensor data, it is possible to use semantic grid maps for localization.

The use of semantic information has additional benefits. This information can be obtained from different sensors which allows using expensive and precise sensors for mapping, and low-cost sensors for localization. Moreover, once extracted, semantic information is invariant to illumination and weather conditions. The ability to distinguish different objects can also be helpful for other tasks, such as navigation or movable objects (vehicles and pedestrians) tracking.

Localization errors for logs recorded at night are consistent with the ones observed in daylight. Occupancy and reflectivity grid maps led to smaller errors than color or semantic grid maps. This result is expected since LiDAR data are nearly invariant to illumination. The localization with semantic grid maps also managed to maintain position tracking, but with errors higher than the ones found in the daylight logs. This is also expected since the method relies on images that are more impacted by illumination conditions.

Experiments with the logs UFES_TD and PARK_TD show that the localization with occupancy, reflectivity, or semantic grid maps is robust to (i) distractors (e.g., parked cars and pedestrians), (ii) changes in the environment due to time passing, (iii) different conditions of illumination, and (iv) presence of U-turns.

In log URBAN_TD and DIVERSE configuration, the localization with occupancy grid maps diverged from the true path. By analyzing the log, we found that the divergence happened in a situation of intense traffic in which the self-driving car was surrounded by other vehicles as exemplified by Figure 14. The divergence was not observed when using STABLE configuration.

By adding less noise into the particles, the particle spreading slowed down, allowing surrounding cars to move before the localization drifted to an unrecoverable state.

In log AIRPRT_TM, the localization based on occupancy and reflectivity grid maps also achieved higher performance. The high localization accuracy is surprising in this log given the apparent lack of features in the environment. In particular, lane marks are not visible in occupancy grid maps and the localization with this map relies mostly on the borders of the runway and on vegetation for constraining the car pose. This was the log in which the localization based on semantic grid maps achieved the highest errors.

Both, the localization based on color and semantic grid maps drifted from the correct poses at the beginning of the log, and the offset was maintained until the end. The poor illumination conditions and the fact that the environment is significantly different from the ones observed in the CityScapes dataset may explain the high errors when using semantic grid maps.

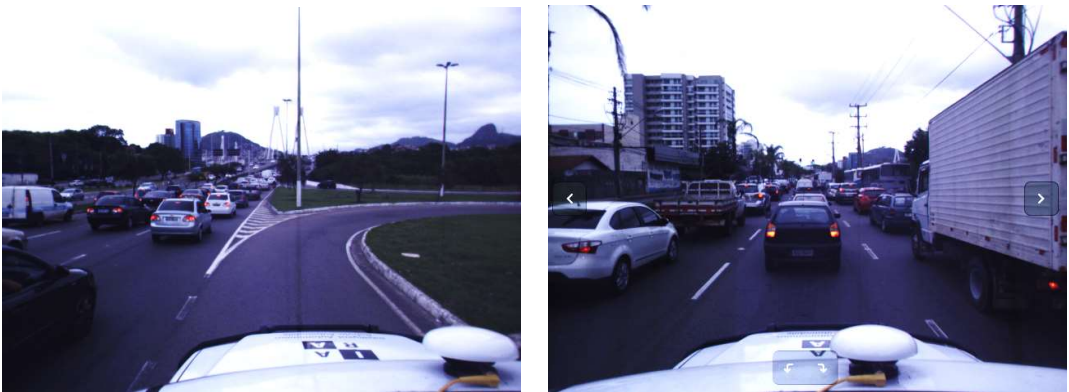

Figure 14. Intense traffic conditions present in the URBAN dataset.

## VII. Conclusion

This work compared the accuracy of a localization system for self-driving cars using different types of grid maps. For that, novel techniques for building grid maps of complex large-scale environments, and for estimating the localization ground truth were also proposed. Experiments were carried out in diverse environments under challenging conditions. The mapping technique was successfully employed for building occupancy, reflectivity, color, and semantic grid maps of all environments. Experimental results showed that occupancy and reflectivity grid maps, and the DIVERSE configuration of parameters achieved more accurate results. Semantic grid maps also led to reasonable accuracy demonstrating that this is a promising area of research. These maps contain features that are helpful for other tasks such as planning which motivates further their use. On the other hand, color grid maps produced inaccurate localization even using the robust ECC measure for computing particles weights. Additional studies are necessary to find out how to effectively use color. Such studies are worth mainly because cameras have lower costs than LiDARs, and since humans can drive relying solely on vision, the localization with color grid maps is likely to be possible.

Besides varying illumination conditions, the inferior performance of semantic and color grid maps in comparison to maps based only on LiDAR can be explained by errors in the fusion of Velodyne and camera. Among others, lack of synchrony between sensors, latency in camera (particularly in low light situations), and inaccurate calibration of sensors' relative poses are potential sources of errors. The semantic segmentation of point clouds is a potential way of preventing these issues.

In future works, we intend to evaluate the localization in different types of environments (e.g., highways and hills), conditions of operation (e.g., diverse weather), and with other grid maps found in literature (e.g., height grid maps (Wolcott & Eustice, 2017)). In addition, we will study the impact of using two or more maps for localization. Since maps may contain complementary information, together they can overcome limitations found in each individual one. As an example, in our experiments the localization with occupancy grid maps diverged when the self-driving car was surrounded by other vehicles. This divergence could possibly be prevented by using additional information besides occupancy probabilities. Unfortunately, using more data increases computational requirements. Therefore, future studies should seek the set of information that is not only sufficient for accurate localization, but also that is the minimum necessary.

### ACKNOWLEDGMENTS

This study was financed in part by the Coordenação de Aperfeiçoamento de Pessoal de Nível Superior - Brasil (CAPES) - Finance Code 001, Conselho Nacional de Desenvolvimento Científico e Tecnológico (CNPq, Brazil) and Fundação de Amparo à Pesquisa do Espírito Santo - Brasil (FAPES) – grant 84412844. We thank NVIDIA for providing GPUs used in this research. Filipe Mutz thanks the Instituto Federal do Espírito Santo (IFES) for encouraging this research through the Productividade Researcher Program (Programa Pesquisador Produtividade – PPP).